\documentclass[a4paper]{article}

\usepackage{INTERSPEECH_v2}
\usepackage{empheq} 
\usepackage{multirow}
\usepackage{slashbox}
\usepackage{url}

\title{Improving speech recognition by revising gated recurrent units}
\name{Mirco Ravanelli$^1$, Philemon Brakel$^2$, Maurizio Omologo$^1$, Yoshua Bengio$^2$}
\address{
  $^1$Fondazione Bruno Kessler, Trento, Italy\\
  $^2$Universit\'e de Montr\'eal, Montr\'eal, Canada}
\email{mravanelli@fbk.eu, pbpop3@gmail.com, omologo@fbk.eu, yoshua.umontreal@gmail.com}

\begin{document}

\maketitle
\begin{abstract}
Speech recognition is largely taking advantage of deep learning, showing that substantial benefits can be obtained by modern Recurrent Neural Networks (RNNs). The most popular RNNs are Long Short-Term Memory (LSTMs), which typically reach state-of-the-art performance in many tasks thanks to their ability to learn long-term dependencies and robustness to vanishing gradients. Nevertheless, LSTMs have a rather complex design with three multiplicative gates, that might impair their efficient implementation. An attempt to simplify LSTMs has recently led to Gated Recurrent Units (GRUs), which are based on just two multiplicative gates.

This paper builds on these efforts by further revising GRUs and proposing a simplified architecture potentially more suitable for speech recognition. The contribution of this work is two-fold. First, we suggest to remove the reset gate in the GRU design, resulting in a more efficient single-gate architecture. Second, we propose to replace tanh with ReLU activations in the state update equations. Results show that, in our implementation, the revised architecture reduces the per-epoch training time with more than 30\% and consistently improves recognition performance across different tasks, input features, and noisy conditions when compared to a standard GRU.

\end{abstract}

\section{Introduction}
Building machines that are able to recognize speech represents a fundamental step towards flexible and natural human-machine interfaces. A primary role for improving such a technology is played by deep learning \cite{Goodfellow-et-al-2016-Book}, which has recently contributed to significantly outperform previous GMM/HMM speech recognizers \cite{lideng}. 
During the last years, deep learning has been rapidly evolving, progressively offering more powerful and robust techniques, including effective regularization methods \cite{dropout,batchnorm}, improved optimization algorithms \cite{adam}, as well as better network architectures.
The early deep learning works in speech recognition were mainly based on standard multilayer perceptrons (MLPs) \cite{IEEEexample:intro7,IEEEexample:intro1}, while recent systems benefit from more advanced architectures, such as Convolutional Neural Networks (CNNs) \cite{cnn1} and Time Delay Neural Network (TDNN) \cite{tdnn,tdnn2}.

Nevertheless, since speech is inherently a sequential signal, it would be natural to address it with Recurrent Neural Networks (RNNs), which are potentially able to properly capture long-term dependencies. 
Several works have already highlighted the effectiveness of RNNs in various speech processing tasks such as speech recognition \cite{graves,lstm_speech,baidu,joint6,chime4_paper}, speech enhancement \cite{dnn_se3}, speech separation \cite{sep_lstm,ndnn1} as well as speech activity detection \cite{lstm_vad}. Recent results in the newborn field of end-to-end speech recognition \cite{e2e1,attention1} have also shown that RNNs are promising candidates for replacing traditional Hidden Markov Models (HMMs) in DNN/HMM speech recognizers.

Training RNNs, however, can be complicated by vanishing and exploding gradients, which might impair learning long-term dependencies \cite{Bengio94}.
Although exploding gradients can effectively be tackled with simple clipping strategies \cite{pascanau}, the vanishing gradient problem requires special architectures to be properly addressed. A common approach relies on 
the so-called gated RNNs, whose core idea is to introduce a gating mechanism for better controlling the flow of the information through the various time-steps. Within this family of architectures, vanishing gradient issues are mitigated by creating effective ``shortcuts", in which the gradients can  bypass multiple temporal steps.

The most popular gated RNNs are Long Short-Term Memory networks (LSTMs) \cite{lstm}, which often achieve state-of-the-art performance in several machine learning tasks. 
 LSTMs rely on a network design consisting of memory cells which are controlled by forget, input, and output gates.
Despite their effectiveness, such a sophisticated gating mechanism might result in an overly complex model that can be tricky to implement efficiently. On the other hand, computational efficiency is a crucial issue for RNNs and considerable research efforts have recently been devoted to the development of alternative architectures \cite{lstm_odyssey,gru3,lstm_highway}.
With this purpose, an attempt to simplify LSTMs has led to a novel model called Gated Recurrent Unit (GRU) \cite{gru1,gru2}, which is based on just two multiplicative gates. Despite the adoption of a simplified gating mechanism, some works in the literature agree that GRUs and LSTMs provide a comparable performance in different machine learning tasks \cite{gru2,gru3,gru4}.

This work continues these efforts by further revising GRUs and proposing an architecture potentially more suitable for speech recognition. The contribution of this paper is twofold: First, we propose to remove the reset gate from the network design. Similarly to \cite{mgru}, we found that removing the reset gate does not affect the system performance, since we observed a certain redundancy in the role played by the update and reset gates. 
Second, we propose to replace hyperbolic tangent (tanh) with Rectified Linear Units (ReLU) activations in the state update equation. In the past, such a non-linearity has been avoided for RNNs due to the numerical instabilities caused by the unboundedness of ReLU activations. However, when coupling our ReLU-based GRU architecture with batch normalization \cite{batchnorm}, we did not experience such numerical issues. This allows us to take advantage of ReLU neurons, which have been proven effective at further alleviating the vanishing gradient problem as well as speeding up network training.    

Our results, obtained on different tasks, input features, and noisy conditions  show that, in our implementation, the revised architecture reduces the per-epoch training wall-clock time with more than 30\%, while improving the recognition performance over all the experimental conditions considered in this work.  

\section{Revising GRUs} \label{sec:rev_gru}
The GRU architecture is defined by the following equations: 

\begin{subequations}
\begin{align}
z_{t}&=\sigma(W_{z}x_{t}+U_{z}h_{t-1}+b_{z}), \\
\label{eq:eq_2}r_{t}&=\sigma(W_{r}x_{t}+U_{r}h_{t-1}+b_{r}), \\
\label{eq:eq_3}\widetilde{h_{t}}&=\tanh(W_{h}x_{t}+U_{h}(h_{t-1}r_{t})+b_{h}), \\
\label{eq:eq_4}h_{t}&=z_{t}h_{t-1}+ (1-z_{t}) \widetilde{h_{t}}.
\end{align}
\end{subequations}

In particular, $z_{t}$ and $r_{t}$ are vectors corresponding to the update and reset gates respectively, while $h_{t}$ represents the state vector for the current time frame $t$.
The activations of both gates are element-wise logistic sigmoid functions $\sigma(\cdot)$, which constrain $z_{t}$ and $r_{t}$ to take values ranging from 0 and 1. The candidate state $\widetilde{h_{t}}$ is processed with a hyperbolic tangent. 
The network is fed by the current input vector $x_{t}$ (e.g., a vector with speech features), while the parameters of the model are the matrices $W_z$, $W_r$, $W_h$ (the feed-forward connections) and $U_z$, $U_r$, $U_h$ (the recurrent weights).
The architecture finally includes trainable bias vectors $b_z$, $b_r$ and $b_h$, which are added before the non linearities. 

As shown in Eq.~\ref{eq:eq_4}, the current state vector $h_{t}$ is a linear interpolation between the previous activation $h_{t-1}$ and the current candidate state $\widetilde{h_{t}}$. The weighting factors are set by the update gate $z_{t}$, that decides how much the units will update their activations. Note that such a linear interpolation, which is similar to that adopted for LSTMs \cite{lstm}, is the key component for learning long-term dependencies. For instance, if $z_{t}$ is close to one, the previous state is kept unaltered and can remain unchanged for an arbitrary number of time steps. On the other hand, if $z_{t}$ is close to zero, the network tends to favor the candidate state $\widetilde{h_{t}}$, which depends more heavily on the current input and on the closer hidden states. The candidate state $\widetilde{h_{t}}$, also depends on the reset gate $r_{t}$, which allows the model to delete the past memory by forgetting the previously computed states. 




The model proposed in this paper is a revised version of the GRUs described above. The main changes concern reset gate and ReLU activations, as outlined in the next sub-sections. 

\subsection{Removing the reset gate}
The reset gate can be useful when significant discontinuities occur in the sequence. For language modeling, this may happen when moving from one text to another one which is not-semantically related. In such situations, it is convenient to reset the stored memory in order to avoid taking a decision biased by an uncorrelated history. However, for some specific tasks such a functionality might not be useful. In \cite{mgru}, for instance, removing $r_{t}$ from the GRU model has led to a single-gate architecture called Minimal Gated Recurrent Unit (M-GRU), which achieves a performance comparable to that obtained by standard GRUs in handwritten digit recognition as well as in a sentiment classification task.  

We argue that the role played by the reset gate should be reconsidered also for acoustic modeling in speech recognition. In fact, a speech signal is a sequence that evolves rather slowly (the features are typically computed every 10 ms), in which the past history can virtually always be helpful.  
Even in the presence of strong discontinuities, for instance observable at the boundary between a vowel and a fricative, completely resetting the past memory can be harmful. On the other hand, it is helpful to memorize phonotactic features, since some phone transitions are more likely than others.

\begin{figure}
\centering
\includegraphics[width=0.45\textwidth]{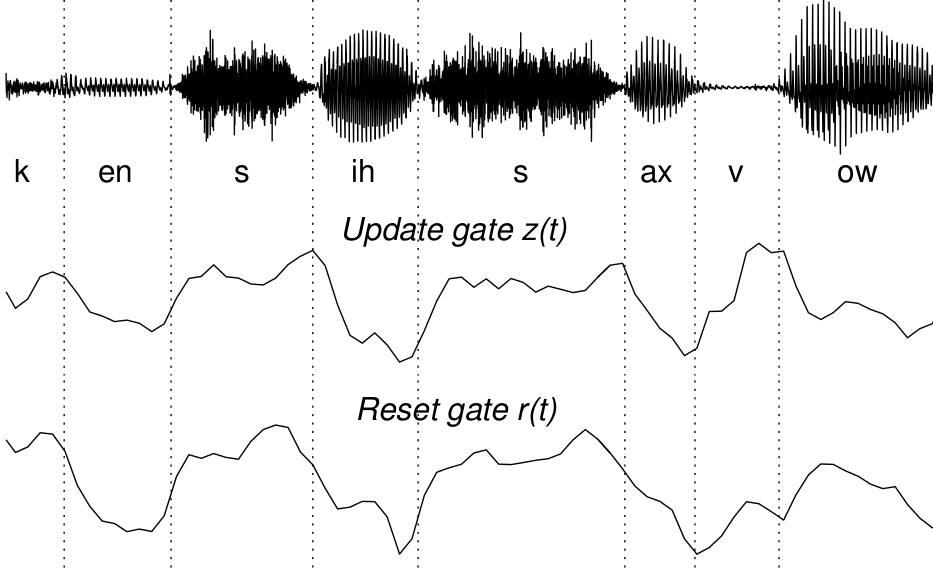}
\caption{Average activations of update and reset gates for a GRU trained on TIMIT in a chunk of the utterance ``sx403" of speaker ``faks0".}
\label{fig:im1}
\end{figure}

Moreover, we believe that a certain redundancy in the activations of reset and update gates might occur when processing speech sequences.  For instance, when it is necessary to give more importance to the current information,  the GRU model can set small values of $r_{t}$. A similar effect can also be achieved with the update gate only, by setting small values of $z_{t}$. The latter solution tends to weight more the candidate state $\widetilde{h_{t}}$, which, as desired, depends more heavily on the current input and on its more recent history.  
Similarly, 
a high value can be assigned either to $r_{t}$ or to $z_{t}$, in order to place more importance on past states. 
This redundancy is also highlighted in Fig. \ref{fig:im1}, where a temporal correlation in the average activations of update and reset gates 
can be readily appreciated for a GRU trained on TIMIT. 


The first modification to standard GRUs proposed in this work thus concerns the removal of the reset gate $r_{t}$, which helps in limiting the redundancy in the gating mechanism. 
The main benefits of this intervention are related to the improved computational efficiency, which is achieved thanks to the reduced number of parameters necessary to reach the performance of a standard GRU.

\subsection{ReLU activations}
The second modification consists in replacing the standard hyperbolic tangent with ReLU activations in the state update equations (Eq.~\ref{eq:eq_3} -~\ref{eq:eq_4}). Tanh activations are, indeed, rather critical since their saturation slows down the training process and causes vanishing gradient issues. The adoption of ReLU-based neurons, which have shown effective in improving such limitations, was not so common in the past for RNNs, due to numerical instabilities originated by the unbounded ReLU functions applied over long time series. Nevertheless, some recent works have shown that ReLU RNNs can be effectively trained with a proper orthogonal initialization \cite{orth_init}. 

Formally, removing the reset gate and replacing the hyperbolic tangent function with the ReLU activation now leads to the following update equations:
\begin{subequations}
\begin{align}
\label{eq:eq_2a}&z_{t}=\sigma(W_{z}x_{t}+U_{z}h_{t-1}+b_{z}), \\
&\widetilde{h_{t}}=\mbox{ReLU}(W_{h}x_{t}+U_{h}h_{t-1}+b_{h}), \\
\label{eq:eq_2c}&h_{t}=z_{t}h_{t-1}+ (1-z_{t}) \widetilde{h_{t}}.
\end{align}
\end{subequations}
We called this architecture M-reluGRU, to emphasize the modifications carried out on a standard GRU.

\subsection{Batch Normalization}
Batch normalization \cite{batchnorm} has been recently proposed in the machine learning community and addresses the so-called \textit{internal covariate shift} problem by normalizing, for each training mini-batch, the mean and the variance of each layer pre-activations. Such a technique has shown to be crucial both to improve the system performance and to speed-up the training procedure. Batch normalization can be applied to RNNs in different ways. In \cite{cesar}, authors suggest to apply it to feed-forward connections only, while in \cite{tim} the normalization step is extended to recurrent connections, using separate statistics for each time-step. 

In this work, we tried both approaches and we observed a comparable performance between them.
We also noticed that coupling the proposed model with batch-normalization \cite{batchnorm} helps in avoiding the numerical issues that often occur when dealing with ReLU RNNs applied to long time sequences. Batch normalization, indeed, rescales the neuron pre-activations, inherently bounding the values of the ReLU neurons.
This allows the network to take advantage of the well-known benefits of such activations.

\section{Experimental setup}
\subsection{Corpora and tasks}
\label{sec:corpora}
To provide an accurate evaluation of the proposed architecture, the experimental validation was conducted using different training datasets, tasks and environmental conditions. 
A first set of experiments with TIMIT was performed to test the proposed model in a close-talking scenario. The experiments with TIMIT are based on the standard phoneme recognition task, which is aligned with that proposed in the Kaldi s5 recipe \cite{kaldi_short}.

A set of experiments was also conducted in a distant-talking scenario with a Wall Street Journal (WSJ) task, to validate our model in a more realistic situation. The reference context was a domestic environment characterized by the presence of non-stationary noise (with an average SNR of about 10dB) and acoustic reverberation (with an average reverberation time $T_{60}$ of about 0.7 seconds).
The original WSJ training dataset was contaminated with a set of impulse responses measured in a real apartment.
The test phase was carried out with the DIRHA-English corpus\footnote{This dataset is being distributed by the Linguistic Data Consortium (LDC).} (with both real and simulated datasets), consisting of 409 WSJ sentences uttered by six native American speakers in the same apartment. A development set of 310 WSJ sentences uttered by six different speakers was also used for hyperparameter tuning.
More details on this dataset and on the impulse responses can be found in \cite{dirha_asru,rav_is16}.

\subsection{System details}

The architecture adopted for the experiments consisted of multiple recurrent layers, which were stacked together prior to the final softmax context-dependent classifier. These recurrent layers were bidirectional RNNs \cite{graves}, which were obtained by concatenating the forward hidden states (collected by processing the sequence from the beginning to the end) with backward hidden states (gathered by scanning the speech in the reverse time order).
Recurrent dropout was used as regularization technique. Since extending standard dropout to recurrent connections hinders learning long-term dependencies, we followed the approach introduced in \cite{drop_asru}, which tackles this issue by sharing the same dropout mask across all the time steps. Moreover,  batch normalization was adopted exploiting the method suggested in \cite{cesar}, as discussed in Sec.~\ref{sec:rev_gru}.
The feed-forward connections of the architecture were initialized according to the \textit{Glorot} initialization \cite{xavier}, while recurrent weights were initialized with orthogonal initialization \cite{orth_init}. Similarly to \cite{ravanelli_SLT}, the gain factor $\gamma$ of batch normalization was initialized to $\gamma=0.1$ and the shift parameter $\beta$ was initialized to 0.

Before training, the sentences were sorted in ascending order according to their lengths and, starting from the shortest utterances, minibatches of 8 sentences were progressively processed by the training algorithm. 
This sorting approach, besides minimizing zero-paddings when forming mini-batches, exploits a curriculum learning strategy \cite{curriculum} which has been shown to improve the performance and to ensure numerical stability of gradients. The optimization was done using the Adam algorithm \cite{adam}, which ran for 22 epochs. The performance on the development set was monitored after each epoch, while the learning rate was halved when the performance improvement went below a certain threshold. Gradient truncation was not applied, allowing the system to learn arbitrarily long time dependences.
The speech recognition labels were derived by performing a forced alignment procedure on the original training datasets. See the standard s5 recipe of Kaldi for more details \cite{kaldi_short}.

To evaluate the proposed architecture on different input features, three sets of experiments were performed with 39 MFCCs, with 40 mel-filterbank coefficients, as well as with 40 fMLLR features derived by a Speaker Adaptive Training (SAT) procedure \cite{kaldi_short}. 
All of these feature vectors were computed every 10 ms with a frame length of 25 ms. 

The main hyperparameters of the model (i.e., learning rate, number of hidden layers, hidden neurons per layer, dropout factor) were optimized on the development data. 
In particular, we guessed some initial values according to our experience, and starting from them we performed a grid search to progressively explore better configurations. A total of 20-25 experiments were conducted for the RNN models.
As a result, an initial learning rate of 0.0013 and a dropout factor of 0.2 were chosen for all the experiments. The optimal numbers of hidden layers and hidden neurons, instead, depend on the considered dataset/model, and range from 4 to 5 hidden layers with 375-607 neurons.

The proposed system was implemented with Theano \cite{theano} and coupled with the Kaldi decoder \cite{kaldi_short} to form a context-dependent DNN/HMM speech recognizer\footnote{The code used for the experiments can be found at \url{https://github.com/mravanelli/theano-kaldi-rnn/}.}.

\section{Results}
In the following sub-sections, a comparison of the proposed architecture with other popular RNNs is presented. The results will be reported for the standard close-talking TIMIT dataset and for the distant-talking WSJ task based on the DIRHA English dataset.
\subsection{Results on TIMIT}

Table \ref{tab:res1} presents the results obtained with the TIMIT dataset. To perform a more accurate comparison of the various architectures, at least five experiments varying the initialization seeds were conducted for each RNN model. The results in Table \ref{tab:res1} are reported as the average Phone Error Rates (PER\%) with their corresponding standard deviations. 

\begin{table}[t!]
\centering
\tabcolsep=0.20cm
    \begin{tabular}{ | l | c | c | c | c | }
    \cline{1-4}
   {\backslashbox{\em{Arch.}}{\em{Feat.}}} & MFCC &  FBANK & fMLLR \\ \hline
relu-RNN & 18.7 $\pm$ 0.18 & 18.3 $\pm$ 0.23 & 16.3  $\pm$ 0.11 \\ \hline
LSTM & 18.1 $\pm$ 0.33 & 17.1 $\pm$ 0.36 & 15.7  $\pm$ 0.32 \\ \hline
GRU & 17.1 $\pm$ 0.20 & 16.7 $\pm$ 0.36 & 15.3  $\pm$ 0.28 \\ \hline
M-GRU & 17.2 $\pm$ 0.11 & 16.7 $\pm$ 0.19 & 15.2  $\pm$ 0.10 \\ \hline
M-reluGRU & \textbf{16.7} $\pm$ 0.26 & \textbf{15.8} $\pm$ 0.10 & \textbf{14.9}  $\pm$ 0.27
\\ \hline  
    \end{tabular}
\caption{Phone Error Rate (PER\%) obtained for the test set of TIMIT with various RNN architectures.}
\label{tab:res1}
\end{table}

\begin{table}[t!]
\centering
\tabcolsep=0.12cm
    \begin{tabular}{  | l | c | c | c | c | c |}
    \cline{1-5}
Architecture & Layers & Neurons & \# Params &  Training time \\ \hline
relu-RNN & 4 & 607 & 6.1 M & 318 sec.  \\ \hline
LSTM & 5 & 375 & 8.8 M & 506 sec.  \\ \hline
GRU & 5 & 465  & 10.3 M & 580 sec. \\ \hline
M-GRU & 5 & 465 & 7.4 M & 390 sec. \\ \hline
M-reluGRU & 5 & 465 & 7.4 M & 390 sec. \\ \hline  
    \end{tabular}
\caption{Comparison of the per-epoch training time for the RNN architectures optimized on the TIMIT development set.}
\label{tab:time}
\end{table}

The first row of Table \ref{tab:res1} presents the results of a traditional RNN with ReLU activations (no gating mechanisms are used here). Although this architecture has recently shown promising results in some machine learning tasks \cite{orth_init}, our speech recognition performance results confirm that gated recurrent networks (rows 2-5) still outperform traditional RNNs.
We also observed that GRUs tend to slightly outperform the LSTM, in the tasks addressed in these experiments.

The M-GRU architecture is the version of GRU without the reset gate. Table \ref{tab:res1} highlights that the M-GRU achieves a performance very similar to that obtained with standard GRUs, further confirming our speculation on the negligible role played by the reset gate in a speech recognition application. 
The last row of Table \ref{tab:res1} reports the performance achieved with the proposed model, in which, besides removing the reset gate, ReLU activations are used. 
Results indicate that M-reluGRU consistently achieves the best performance across the various input features. A remarkable achievement is the average PER(\%) of $14.9$\% obtained with the proposed architecture using fMLLR features. To the best of our knowledge, such a result yields the best published performance on the TIMIT test-set.

Table ~\ref{tab:time} reports a comparison of the per-epoch training time for all the considered RNNs. Results confirm that the main advantage of the proposed model is its ability to significantly reduce the training time. In our Theano implementation, running each GRU epoch lasts 580 seconds on an NVIDIA K40 GPU against just 390 seconds taken by the M-reluGRU, with a training time reduction of more than 32\%.
Table \ref{tab:time} also reports the best architectures obtained after optimizing the hyperparameters on the TIMIT development set. Results show that for GRU models the best performance is achieved with 5 hidden layers of 465 neurons. 

\subsection{Results on DIRHA English WSJ}
Tables \ref{tab:res2} and \ref{tab:res3} summarize the results obtained with the simulated and real parts of the DIRHA English WSJ dataset. For the sake of compactness, only the average performance (obtained by running five experiments with different initialization seeds) is reported, while standard deviations (which range from 0.2\% to 0.4\%) are omitted. 

\begin{table}[t!]
\centering
\tabcolsep=0.24cm
    \begin{tabular}{ | l | c | c | c | c | }
    \cline{1-4}
   {\backslashbox{\em{Arch.}}{\em{Feat.}}} & MFCC &  FBANK & fMLLR \\ \hline
relu-RNN & 23.7  & 23.5 & 18.9 \\ \hline
LSTM & 23.2  & 23.2 & 18.9 \\ \hline
GRU & 22.3  & 22.5 & 18.6 \\ \hline
M-GRU & 21.5  & 22.0 & 18.0 \\ \hline
M-reluGRU & \textbf{21.3}  & \textbf{21.4} & \textbf{17.6} \\ \hline  
\end{tabular}
\caption{Word Error Rate (\%) obtained with the DIRHA English WSJ dataset (simulated part) for various RNN architectures.}
\label{tab:res2}
\end{table}

\begin{table}[t!]
\centering
\tabcolsep=0.24cm
    \begin{tabular}{ | l | c | c | c | c | }
    \cline{1-4}
   {\backslashbox{\em{Arch.}}{\em{Feat.}}} & MFCC &  FBANK & fMLLR \\ \hline
relu-RNN & 29.7  & 30.0 & 24.7 \\ \hline
LSTM & 29.5  & 29.1 & 24.6 \\ \hline
GRU & 28.5  & 28.4 & 24.0 \\ \hline
M-GRU & 28.4  & 28.1 & 23.6 \\ \hline
M-reluGRU & \textbf{27.8}  & \textbf{27.6} & \textbf{22.8} \\ \hline  
\end{tabular}
\caption{Word Error Rate (\%) obtained with the DIRHA English WSJ dataset (real part) for various RNN architectures.}
\label{tab:res3}
\end{table}

Table \ref{tab:res2} and \ref{tab:res3} exhibit a trend comparable to that observed for the TIMIT dataset, confirming the effectiveness of the proposed architecture also in a more realistic and challenging scenario. The results are consistent over both real and simulated data as well as across the different features considered in this study. The reduction of the training time is about 36\% (25 minutes for per-epoch training with the proposed model against 40 minutes spent by the standard GRU).

Moreover, the reset gate removal seems to play a more crucial role in the addressed distant-talking scenario.  If the close-talking results reported in Table \ref{tab:res1} highlight comparable error rates between standard GRU and M-GRU, in the distant-talking case we even observe a performance gain when removing the reset gate. We suppose that this behaviour is due to reverberation, which implicitly introduces redundancy in the signal, due to the multiple delayed replicas of each sample. This results in a forward memory effect that smooths energy envelopes as well as sub-band energy contours. Due to this effect, a reset gate mechanism might become ineffective to forget the past.

\section{Conclusions}
In this paper, we revised standard GRUs for speech recognition purposes. The proposed architecture is a simplified version of a GRU, in which the reset gate is removed and ReLU activations are used. The experiments, conducted on different tasks, features and environmental conditions, have confirmed the effectiveness of the proposed model not only to reduce the computational complexity of our implementation (with a reduction of more than 30\% of the training time over a standard GRU), but also to slightly improve the recognition performance. 

Future efforts will be focused on extending this work to different tasks and larger datasets (e.g., switchboard or LibriSpeech) and to test the revised architecture on modern end-to-end systems (such as CTC and attention-based models). To ensure the practical value of the computational gains of our models, these experiments should be replicated using more optimized implementations.



\bibliographystyle{IEEEtran}
\bibliography{refs}

\end{document}